\documentclass[conference]{IEEEtran}
\IEEEoverridecommandlockouts
\usepackage{cite}
\usepackage{amsmath,amssymb,amsfonts}
\usepackage{algorithmic}
\usepackage{graphicx}
\usepackage{textcomp}
\usepackage{xcolor}
\def\BibTeX{{\rm B\kern-.05em{\sc i\kern-.025em b}\kern-.08em
    T\kern-.1667em\lower.7ex\hbox{E}\kern-.125emX}}

\usepackage{graphicx}
\usepackage{amssymb}
\usepackage{array}
\usepackage{comment}
\usepackage{multirow}
\usepackage{subcaption}
\usepackage{threeparttable,booktabs}
\usepackage{etoolbox}

\renewcommand{\tablename}{Table}
\usepackage[justification=centering]{caption} % Figures caption
\renewcommand{\figurename}{Fig.} % Fig.2 (rather than Figure 2)
\usepackage[export]{adjustbox}
\usepackage{caption, subcaption}

\usepackage[pagebackref=true,breaklinks=true,letterpaper=true,colorlinks,bookmarks=false]{hyperref}
\usepackage{cleveref}

\newcommand*{\upSmallFrown}{\mathbin{\raisebox{0.9ex}{$\smallfrown$}}}
\DeclareMathOperator*{\argmin}{argmin}

\definecolor{applegreen}{rgb}{0.0, 0.5, 0.0}
\definecolor{cadmiumred}{rgb}{0.89, 0.0, 0.13}

\begin{document}

%%%%%%%%% TITLE
\title{Multi-Temporal Convolutions for Human Action Recognition in Videos}

\author{\IEEEauthorblockN{Alexandros Stergiou}
\IEEEauthorblockA{\textit{Department of Information and Computing Sciences} \\
\textit{Utrecht University}\\
Utrecht, The Netherlands \\
a.g.stergiou@uu.nl}
\and
\IEEEauthorblockN{Ronald Poppe}
\IEEEauthorblockA{\textit{Department of Information and Computing Sciences} \\
\textit{Utrecht University}\\
Utrecht, The Netherlands \\
r.w.poppe@uu.nl}
}

\maketitle

%%%%%%%%% ABSTRACT
\begin{abstract}
Effective extraction of temporal patterns is crucial for the recognition of temporally varying actions in video. We argue that the fixed-sized spatio-temporal convolution kernels used in convolutional neural networks (CNNs) can be improved to extract informative motions that are executed at different time scales. To address this challenge, we present a novel spatio-temporal convolution block that is capable of extracting spatio-temporal patterns at multiple temporal resolutions. Our proposed multi-temporal convolution (MTConv) blocks utilize two branches that focus on brief and prolonged spatio-temporal patterns, respectively. The extracted time-varying features are aligned in a third branch, with respect to global motion patterns through recurrent cells. The proposed blocks are lightweight and can be integrated into any 3D-CNN architecture. This introduces a substantial reduction in computational costs. Extensive experiments on Kinetics, Moments in Time and HACS action recognition benchmark datasets demonstrate competitive performance of MTConvs compared to the state-of-the-art with a significantly lower computational footprint\footnotemark.
\end{abstract}

\footnotetext{Our code is available at: \url{https://git.io/JfuPi}}

%%% Introduction
\section{Introduction}
\label{sec:intro}

\begin{figure}[t]
\includegraphics[width=\linewidth]{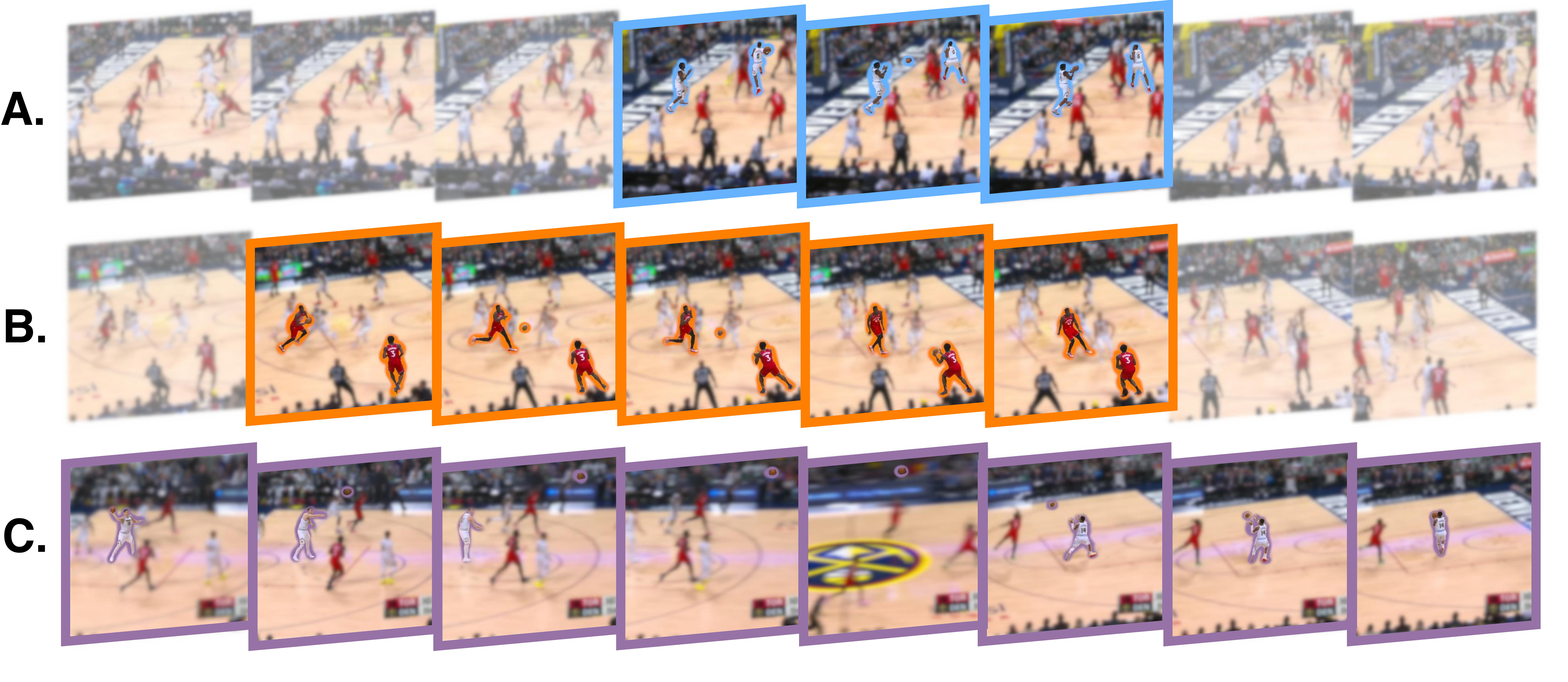}
\caption{Three examples of basketball passes of different duration: (A) brief hand-off pass, (B) longer wing pass, and (C) full-court pass spanning the entire clip duration.}
\label{fig:landing_figure}
\end{figure}

% Intro for action recognition
The variations in how humans execute tasks and how they interact with each other present significant challenges for the recognition of their actions in videos \cite{stergiou2019analyzing}. Differences in visual appearance can largely be captured by deep convolutional neural networks (CNNs). For action recognition in videos, 2D convolutions have been successfully extended to 3D convolutions to additionally extract informative patterns over time. Although connections between the three dimensions do exist \cite{dong1995statistics}, the symmetrical processing of temporal and spatial information significantly limits how the variations in the execution of actions over time can be modeled. For example, as shown in \Cref{fig:landing_figure}, examples in the same action category can significantly differ in the type and duration of the performed movements.

% 3D Convolutions vs time variations
Current efforts in action recognition are based on 3D convolutions \cite{ji20133d} with fixed-sized kernels. We observe that variations in the duration of the performance of an action are typically not in the order of magnitude. Therefore, we argue that we can capture differences in temporal movements by extracting spatio-temporal patterns at different timescales. This approach can flexibly deal with temporal variations while maintaining the spatial modeling power of the method unaffected.

\begin{figure*}[ht]
\includegraphics[width=\textwidth]{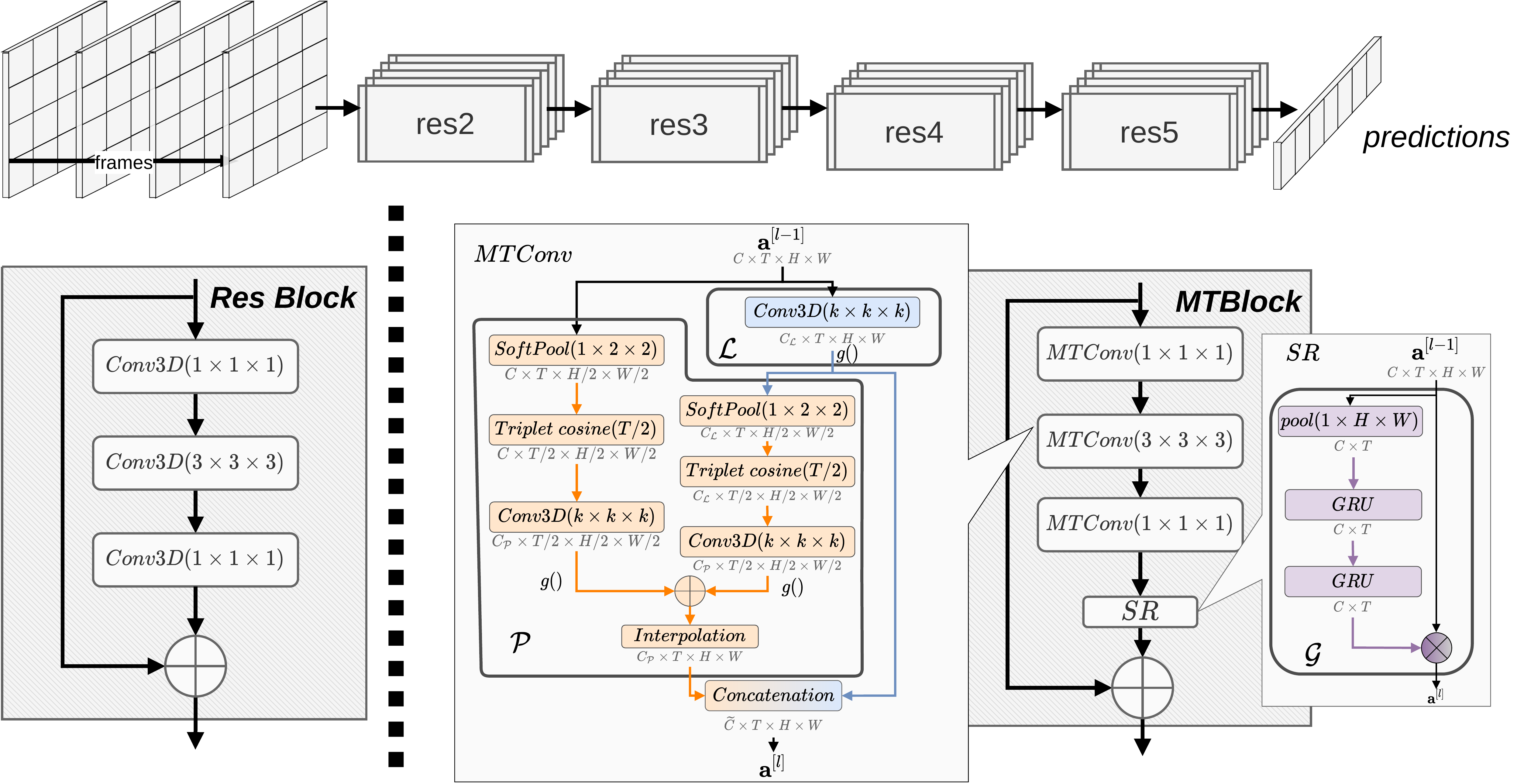}
\caption{\textbf{MTNet architecture}. With X3D~\cite{feichtenhofer2020x3d} ResNet as backbone (top), the proposed MTBlocks (right) are used as direct replacements for Residual blocks (Res Block, left). Blocks contain three consecutive multi-temporal convolutions (MTConv, center) followed by the Squeeze and Recursion (SR, far right) feature alignment. We denote element-wise additions with $\oplus$ and element-wise multiplications with $\otimes$.}
\label{fig:mtconv}
\end{figure*}

% Our focus
In this work, we introduce convolutional blocks that can model temporally variant motion patterns and their cross-feature dependencies. We observe that contributing factors relating to the temporal complexity of actions are primarily data-driven, such as camera motion and video frame rates, or performance-related corresponding to the actor's \textit{prepotent identity} of an action \cite{vallacher2011action}. Motivated by this distinction, we believe that 3D convolutions are inherently constrained to capture only fixed-sized local patterns. To address the temporal complexity of human actions in videos, we propose a multi-temporal convolution (MTConv) block that captures spatio-temporal features variations within their representations. As shown in \Cref{fig:mtconv}, the blocks consist of three branches: a \textit{local branch} ($\mathcal{L}$), a \textit{prolonged branch} ($\mathcal{P}$), and a \textit{global aggregated feature importance branch}. The local branch and prolonged branches focus on spatio-temporal patterns that are performed in short and longer spatio-temporal windows, respectively. The global aggregated feature importance branch aligns the activations of these two branches based the temporal dynamics of motions across the entire video. The novel design of MTConvs, and the resulting MTBlocks, enable the discovery of patterns across time-scales as well as their combination of the learned temporal attention over the entire video sequence. Owing to this property, we can capture local feature dependencies within the (global) scale of the entire sequence.

% evaluation
We validate the proposed blocks on action classification for 3rd person videos, on four large-scale benchmark datasets and an additional fine-tuning dataset. We report performance on Kinetics-400 \cite{carreira2017quo} demonstrating the descriptive quality of MTConv features over equivalent state-of-the-art approaches with the same and larger computational requirements. We further test our method on the Moments in Time \cite{monfort2018moments} dataset that includes large motion variances across examples of the same action class. Equivalently, we present results on the recent Kinetics-700 \cite{carreira2019short} and HACS \cite{zhao2019hacs} datasets. The feature transfer capabilities are then tested with the pre-trained models used on UCF-101 \cite{soomro2012ucf101}.

% paper outline
We discuss current progress in action recognition in \Cref{sec:related}. We then introduce the proposed MTConv blocks in \Cref{sec:method}. Evaluation of our work is shown in \Cref{sec:results}, and we conclude in \Cref{sec:conclusions}.

\section{Related work}
\label{sec:related}

% Two-steams with optical flow
\textbf{Two-stream networks}. Optical flow is a widely used method for representing motion information across video frames. In two-stream networks \cite{simonyan2014two}, one stream is responsible for handling individual (RGB) frames sequentially, while the other processes the optical flow equivalent motions. Information from both streams is then fused at the end. Enhancements to the two-stream approach include the addition of lateral connections between the two streams, for the inclusion of temporal information within spatial patterns \cite{feichtenhofer2016spatiotemporal}, division into temporal segments \cite{wang2016temporal} and spatial-based and temporal-based encoding of segments \cite{diba2017deep}. The main limitation of the two-stream approach is the strong dependence on hand-coded optical flow inputs that prevents the joint learning of complex spatio-temporal features in an end-to-end manner.

% 3D Convolutions
\textbf{3D convolutions}. An alternative approach for utilizing temporal information is the use of 3D convolutions that operate over space-time volumes of stacked frames \cite{baccouche2011sequential}. The use of 3D convolutions has shown improvements on modeling complex spatio-temporal features \cite{hara2018can,kataoka2020would}. Compared to their 2D counterparts, the large computational requirements of 3D convolutions is a disadvantage with much of the recent literature aimed at improving the efficiency of 3D-CNNs. Tran \textit{et al.} \cite{tran2018closer} have considered decoupling 3D convolutions into temporal-only and spatial-only, thus strongly reducing the number of trainable parameters. 3D grouped convolutions \cite{chen2018multifiber,tran2019video} have also shown benefits on performing convolutions in groups and decreasing the number of GFLOPs without a reduction in accuracy. Other works have focused on step-wise progressive network expansions \cite{feichtenhofer2020x3d} that improved the overall efficiency 3D CNNs. Lin \textit{et al.} \cite{lin2019tsm} have introduced temporal shifting activations, emulating the effect of 3D convolutions while relying on frame-based 2D convolutions. This temporal shift module (TSM) has recently been combined with a 3D convolution method to enable or disable shifting \cite{sudhakaran2020gate}.

% Spatio-temporal stream-based approaches
\textbf{Temporal streams in 3D convolutions}. Inspired by the two-steam 2D CNNs with optical flow, initial works on stream-based 3D CNNs have focused on using dual inputs of stacked RGB frames alongside optical flow \cite{carreira2017quo}. Later works of Feichenhofer \textit{et al.} \cite{feichtenhofer2019slowfast} proposed a 3D model of dual adjacent input-based frame sampling of slow and fast frame rates. With the same rationale, Qiu \textit{et al.} \cite{qiu2019learning} showed how global paths using entire videos as inputs and local paths with local spatio-temporal segments can be used in two separate network pathways. Others have considered block-based approaches with octave convolutions \cite{chen2019drop} to model temporal variation in the frequency domain.

% Our method - Short and Long kernels
Although these methods have shown great promise in extracting robust spatio-temporal features, they do not directly address the complex motion features and their relationships across varying temporal scales. In contrast, we design our tri-branch method to address within convolutional blocks the temporal disparities of actions through the extraction of different periodic spatio-temporal features and the discovery of their dynamics.

% Attention in 3D CNNs
\textbf{Spatio-temporal attention}. Image-based methods \textit{Squeeze and Excitation} \cite{hu2018squeeze} and \textit{Gather and Excite} \cite{hu2018gather} consider self-attention mechanisms as calibration methods for convolutional features in images. Extensions to video have also been proposed based on attention clustering \cite{long2018attention} or through the division of information in terms of appearance and spatial relations \cite{wang2018appearance}. This has also led to the introduction of recurrent sub-networks to explore the dynamics of extracted patterns \cite{stergiou2020learn}. Recent works (e.g. \cite{liu2021tam}) further utilize self-attention for the creation of local importance maps and location-invariant feature weights. 

% Our method attention
Our method uses temporal attention through the alignment of activations with Squeeze and Recursion modules \cite{stergiou2020learn} on the \textit{global aggregated feature importance} branch. This enforces coherence of the learned spatio-temporal patterns despite temporal variations.

\section{Multi-Temporal Networks}
\label{sec:method}

% Section outline
In this section, we first describe multi-temporal convolutions (MTConvs) and their inner workings in terms of how information is processed. We then detail the structure of the blocks (\textit{MTBlocks}, shown in \Cref{fig:mtconv}) in which they operate in \Cref{sec:method::blocks}. In \Cref{sec:method::network}, we introduce the CNN architecture that employs these MTBlocks.

Formally, layer activations are denoted by $\textbf{a}_{(C \! \times \! T \! \times \! H \! \times \! W)}$ with $C$ channels, $T$ frames, $H$ height and $W$ width. Branch activations are denoted by $\textbf{a}^{\mathcal{L}}$ for the local branch ($\mathcal{L}$) and $\textbf{a}^{\mathcal{P}}$ for the prolonged branch ($\mathcal{P}$). Layers are indexed with $l$ and indicated as $\textbf{a}^{[l]}$ or following $\textbf{a}^{[l],\mathcal{L}}$, $\textbf{a}^{[l],\mathcal{P}}$ in branch notation.

\subsection{Multi-Temporal Convolutions (MTConv)}
\label{sec:method::conv}

% General 
The local and prolonged branches each use a portion of the number of channels ($\widetilde{C}$) of the layer ($l$). To determine the number of channels for each branch, a channel ratio parameter $\delta$ is introduced. We define the channel size of the input activations ($\textbf{a}^{[l-1]}$) as $C$. Channels $C_{\mathcal{L}}$, based on ratio $\delta$, are the lowest integer value approximation (through the homonym function denoted with $\lfloor floor \rfloor$). Respectively, the maximum integer value approximation ($\lceil ceil \rceil$) is used for channels $C_{\mathcal{P}}$:

\begin{equation}
\label{eq:rate}
\begin{split}
    C_{\mathcal{L}} + C_{\mathcal{P}} = \widetilde{C} \; where \qquad \quad\\
    C_{\mathcal{L}} = \lfloor \delta * \widetilde{C}\rfloor \; and \; C_{\mathcal{P}} = \lceil(1-\delta) * \widetilde{C}\rceil \; 
\end{split}
\end{equation}

Inputs are first processed by the $\mathcal{L}$ and $\mathcal{P}$ branches. $\mathcal{L}$ processes a single input $\textbf{a}^{[l-1]}$ of size $(C \! \times \! T \! \times \! H \! \times \! W)$. $\mathcal{P}$ is performed over input pair ($\textbf{a}^{[l],\mathcal{L}}, \textbf{a}^{[l-1]}$) with the first volume being the resulting activations from branch $\mathcal{L}$ of size $(C_{\mathcal{L}} \! \times \! T \! \times \! H \! \times \! W)$ and the second being the original layer input. Dual inputs are used in the prolonged branch ($\mathcal{P}$) as spatio-temporal patterns of elongated duration and spatial sizes are strongly correlated with corresponding more local features, which are extracted by $\mathcal{L}$. With prolonged features incorporating the complexity of local short-term ones, the $\mathcal{P}$ branch effectively operates over $C_{\mathcal{L}}+C_{in}$ channels addressing the added complexity over $\mathcal{L}$. The $\mathcal{L}$ branch and $\mathcal{P}$ branch feature extraction process is summarized as seen in \Cref{eq:form1}:

\begin{equation}
\label{eq:form1}
    \textbf{a}^{[l]} = \mathcal{L}(\textbf{a}^{[l-1]}) \upSmallFrown (\mathcal{P}(\mathcal{L}(\textbf{a}^{[l-1]}), \textbf{a}^{[l-1]}))
\end{equation}

where $\upSmallFrown$ denotes the concatenation of the outputs from the two branches into a single volume.

\textbf{Local branch in MTConv}. The local branch is used for the extraction of short-term local motions within the input activations. Given input ($\textbf{a}^{[l-1]}$) we use a Conv3D followed by batch normalization (BN) \cite{ioffe2015batch} and compute feature volume ($\textbf{z}^{[l],\mathcal{L}}$) of $C_{\mathcal{L}}$ channels followed by non-linearity ($g()$) with ReLU. Unless stated otherwise, $g()$ refers to a ReLU activation. The final branch output takes the form of $\textbf{a}^{[l],\mathcal{L}}=g(\textbf{z}^{[l],\mathcal{L}})$.

\textbf{Prolonged branch in MTConv}. The prolonged branch aims at the extraction of patterns of extended duration , incorporating information from the local branch ($\mathcal{L}$) and the layer input. To explore long-temporal features, both inputs are reduced by a factor of two across their spatio-temporal dimensions. Such a size reduction provides a balanced trade-of between accuracy and computation. More aggressive reduction strategies using larger factors lead to significant information loss. Both inputs are initially downsampled spatially by their per-frame regional exponential maximum with \textit{SoftPool} \cite{stergiou2021refining} with the activations produced being of size $T \! \times \! H' \! \times \! W'$ (where $H'=H/2$ and $W'=W/2$). The activations are then downsampled temporally by a temporal triplet cosine frame selection to size $T' = T/2$. We provide detailed explanations for both methods later in the section. The inclusion of receptive fields twice the duration of those in $\mathcal{L}$ allows for the exploration of temporal movements of larger spatio-temporal regions without the increased computational requirements of kernels double the size. Extended temporal patterns for inputs $\textbf{a}^{[l-1]}$ and $\textbf{a}^{[l],\mathcal{L}}$ are extracted by Conv3D operations followed by BN. The complete process is formulated as follows:

\begin{equation}
\label{eq:prolonged}
    \textbf{a}^{[l],\mathcal{P}} = \mathcal{I}(g(\textbf{z}^{[l],\mathcal{L} \rightarrow \mathcal{P}}) \oplus g(\textbf{z}^{[l],\mathcal{P}}))
\end{equation}

in which, $\oplus$ denotes element-wise addition and $\mathcal{I}()$ is the spatio-temporal tri-linear interpolation of the volume from size ($T' \! \times \! H' \! \times \! W'$) to original size ($T \! \times \! H \! \times \! W$). The feature volume $\textbf{z}^{[l],\mathcal{L} \rightarrow \mathcal{P}}$ corresponds to the extracted patterns from the reduced input $\textbf{a}^{[l],\mathcal{L}}$, while $\textbf{z}^{[l],\mathcal{P}}$ corresponds to features extracted from $\textbf{a}^{[l-1]}$:

\begin{equation}
\label{eq:prolonged_feats}
\centering
    \textbf{z}^{[l],\mathcal{L} \rightarrow \mathcal{P}} = \mathcal{T}(\overline{\textbf{a}}^{[l],\mathcal{L}}) * \textbf{w}^{\mathcal{L} \rightarrow \mathcal{P}} \; and \; \textbf{z}^{[l],\mathcal{P}} = \mathcal{T}(\overline{\textbf{a}}^{[l]}) * \textbf{w}^{\mathcal{P}}
\end{equation}

with $\mathcal{T}()$ being the triplet cosine frame selection for a spatially pooled volume ($\overline{\textbf{a}}$). The convolutional weight vectors for the respective inputs are denoted as $\textbf{w}^{\mathcal{L} \rightarrow \mathcal{P}}$ and $\textbf{w}^{\mathcal{P}}$. 

% Downsampling methods - Spatial
\textbf{Prolonged branch spatial downsampling}. Downsampling blocks use soft-maximum approximation (\textit{SoftPool}, \cite{stergiou2021refining}), to reduce the spatial dimensions of the input activations. The method uses the softmax weights of activations with each of the inputs within the kernel region having a proportional effect to the output. This is formulated given an input $\textbf{a}$ and frame ($t$) region $\textbf{R}$ for size $H \! \times \! W$: 

\begin{equation}
\label{eq:softpool}
\centering
    \overline{\textbf{a}}_{t,r} = \sum_{r \in \textbf{R}}\frac{e^{\textbf{a}_{t,r}} * \textbf{a}_{t,r}}{\sum\limits_{k \in \textbf{R}} e^{\textbf{a}_{t,k}}}, \; \forall \; t \in |T|
\end{equation}

% Downsampling methods - Temporal
\textbf{Prolonged branch temporal downsampling}. The extension of image-based pooling methods to time-inclusive data comes at the expense of a decrease in spatial detail through the fusion of multiple frames. As the proposed method depends on the preservation of such features in order to extract their extended spatio-temporal patterns, we instead introduce a frame-selection sampling method to decrease the temporal dimensionality of the spatially-reduced activation volume ($\overline{\textbf{a}}$). We termed this method \textit{temporal triplet cosine frame selection}.

\begin{figure}[ht]
\centering
\includegraphics[width=\linewidth]{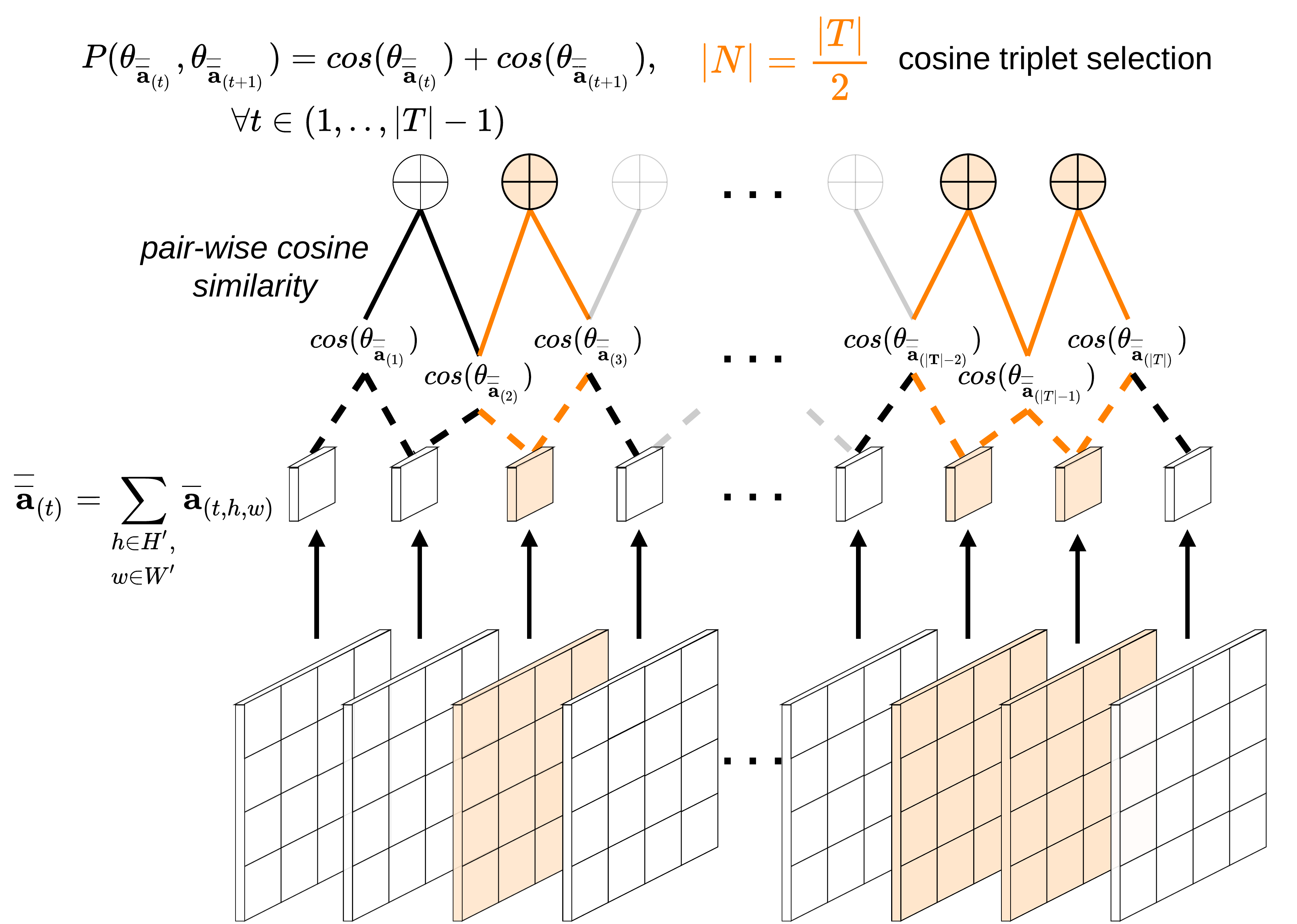}
\caption{\textbf{Temporal triplet cosine similarity frame selection}. Selection is based on the channel-wise sum ($\oplus$) of cosine similarities per pair of adjacent frames ($cos(\theta_{\overline{\overline{\textbf{a}}}_{(t)}})$), calculated from their spatially-summed volumes ($\overline{\overline{\textbf{a}}}$).}
\label{fig:triplet}
\end{figure}

The frame selection process is shown in \Cref{fig:triplet}. Initially, for each frame $t$ all activations are summed spatially to produce a single value per frame. The spatially-summed activation ($\overline{\overline{\textbf{a}}}$) contains frame-wise activation vectors. These vectors can be used to measure the feature-wise similarity per pair of frames using their respective dot product and magnitude:

\begin{equation}
\label{eq:cosine}
\centering
    cos(\theta_{\overline{\overline{\textbf{a}}}_{(t)}}) = \frac{\sum\limits_{c \in C} \overline{\overline{\textbf{a}}}_{(t,c)} * \overline{\overline{\textbf{a}}}_{(t+1,c)}}{\sqrt{\sum\limits_{c \in C}\overline{\overline{\textbf{a}}}_{(t,c)}^{2}} * \sqrt{\sum\limits_{c \in C}\overline{\overline{\textbf{a}}}_{(t+1,c)}^{2}} }
\end{equation}

Their cosine similarity is then summed in similarity pairs ($P(\theta_{\overline{\overline{\textbf{a}}}_{(t)}},\theta_{\overline{\overline{\textbf{a}}}_{(t+1)}}) = cos(\theta_{\overline{\overline{\textbf{a}}}_{(t)}})+cos(\theta_{\overline{\overline{\textbf{a}}}_{(t+1)}})$) for the creation of triplets. This represents a concatenated view of the similarity in features for frame ($t$) in comparison to features of the preceding frame ($t-1$) and succeeding frame ($t+1$). Temporal pooling by triplets then takes the form of selecting the frame locations ($N$) with the lowest $|T|/2$ triplet cosine similarities. This cosine-based sampling can reduce feature redundancy across frames while focusing of frames that are found to be more informative:

\begin{equation}
\label{eq:frame_selection}
\centering
\begin{split}
    \argmin_{\forall n \in N} P(\theta_{\overline{\overline{\textbf{a}}}_{(n)}},\theta_{\overline{\overline{\textbf{a}}}_{(n+1)}}) = cos(\theta_{\overline{\overline{\textbf{a}}}_{(n)}})+cos(\theta_{\overline{\overline{\textbf{a}}}_{(n+1)}}),\\
    \; where \; N \subset T, |N|=|T|/2 \qquad \qquad
\end{split}
\end{equation}

As frames are selected based on their similarity instead of being temporally fused together, the per-frame activations remain consistent over the produced decreased volume.

\subsection{Multi-Temporal Blocks (MTBlocks)}
\label{sec:method::blocks}

% SR
\textbf{Global aggregated feature importance}.We align the concatenated activations of the local ($\mathcal{L}$) and prolonged ($\mathcal{P}$) branches based on the importance of each feature in the context of the entire video sequence. The role of the \textit{global aggregated feature importance} branch ($\mathcal{G}$) is the creation of coherent activations based on averaged feature attention through \textit{Squeeze and Recursion} \cite{stergiou2020learn} with GRU \cite{cho2014learning} recurrent cells. The branch operates over a vectorized version of the original volume pooled by its spatial dimensionality ($pool(\textbf{a}^{[l-1]})$). The pooled volume is processed through a dual-layer recurrent sub-network for the discovery and amplification of globally-informative features. Initial refinement of salient features is done by the update gate ($\textbf{z}_{(t)}$) that uses the per-frame ($t$) instance input ($pool(\textbf{a}^{[l-1]})_{(t)}$) with state ($\textbf{h}_{(t-1)}$) of the previous recurrent cell (for time $t-1$), through a sigmoid ($\sigma$) activation and weight $\textbf{W}_{z}$ (with bias term $\textbf{b}_{z}$):
 
\begin{equation}
\label{eq:update_gate}
    \textbf{z}_{(t)} =\{ \sigma(\textbf{W}_{z} * [\textbf{h}_{(t-1)},pool(\textbf{a}^{[l-1]})_{(t)}] + \textbf{b}_{z}) \} \quad \quad\\
\end{equation}
 
Cell input $pool(\textbf{a}^{[l-1]})_{(t)}$ and previous state outputs $\textbf{h}_{(t-1)}$ also pass through a reset gate ($\textbf{r}_{(t)}$), with weight ($\textbf{W}_{z}$) and bias ($\textbf{b}_{z}$) terms, to ignore temporally inconsistent features. 

\begin{equation}
\label{eq:reset_gate}
    \textbf{r}_{(t)} =\{ \sigma(\textbf{W}_{r} * [\textbf{h}_{(t-1)},pool(\textbf{a}^{[l-1]})_{(t)}] + \textbf{b}_{z}) \} \quad \quad\\
\end{equation}

Both update and reset gates act in a complementary manner on the same inputs. Based on the activations produced by the reset gate, a candidate hidden state is computed ($\widetilde{\textbf{h}}_{(t)}$) with a $tanh$ activation, and reduced influence from the previous state ($\textbf{h}_{(t-1)}$) based on $\textbf{r}_{(t)}$. The produced cell state is the fusion of a proportion of the previous state ($\textbf{z}_{(t)} * \textbf{h}_{(t-1)}$) and the supplementary portion of the candidate hidden state ($(1-\textbf{z}_{(t)}) * \widetilde{\textbf{h}}_{(t)}$), as summarized in \Cref{eq:candidate_out,eq:cell_out}:

\begin{align}
    \widetilde{\textbf{h}}_{(t)} = tanh(\textbf{W}_{h} * [\textbf{r}_{(t)}*\textbf{h}_{(t-1)},pool(\textbf{a}^{[l-1]})_{(t)}] + \textbf{b}_{h}) \label{eq:candidate_out}\\
    \textbf{h}_{(t)} = \textbf{z}_{(t)}*\textbf{h}_{(t-1)}+(1-\textbf{z}_{(t)}) * \widetilde{\textbf{h}}_{(t)} \qquad \quad \label{eq:cell_out}
\end{align}

All cell outputs ($\textbf{h}_{(t)}$) are concatenated to create a filtered activation map of intensities. As shown in \Cref{fig:mtconv}, this temporal excitation volume is used in conjunction with the original input ($\textbf{a}^{[l-1]}$) through an element-wise multiplication operation. The produced activation map ($\textbf{a}^{[l]}$) effectively incorporates the global feature dynamics for the discovered features of different spatio-temporal region sizes.

\subsection{Multi-Temporal Networks (MTNet)} 
\label{sec:method::network}

% General architecture construction
We proposed three MTNet architecture variants that use as backbones the corresponding X3D$_{S}$, X3D$_{M}$ and X3D$_{L}$ models which vary in size and GFLOP usage, and replace their Residual blocks and 3D Convs with the proposed MTBlocks and MTConvs, as shown in \Cref{fig:mtconv}. We denote our models as MTNet$_{S}$, MTNet$_{M}$ and MTNet$_{L}$. The architectures follow a step-wise network and block expansion as recently proposed in video \cite{feichtenhofer2020x3d} and image-based models\cite{radosavovic2020designing}. Details of the three proposed models in terms of the number of parameters and GFLOPs appear in \Cref{table:K400_accuracies}. 

\section{Experiments and Results}
\label{sec:results}

% Section outline
We evaluate our MTBlock and the resulting three MTNets on five popular action recognition benchmark datatsets and compare them against the current state-of-the-art in \Cref{sec:results:sota}. In \Cref{sec:results:ablation}, we compare MTNets with different channel ratios ($\delta$) as well as regular 3D convolutions in terms of classification accuracy and computational complexity. Finally, we evaluate the performance in a transfer learning setting (\Cref{sec:results:TL}).

\subsection{Datasets}
\label{sec:results:datasets}
For our main evaluation, we use four large-scale action recognition datasets. The results presented in this section are calculated on the validation sets of Kinetics-400 (K-400) \cite{carreira2017quo}, the extended Kinetics-700 (K-700) \cite{carreira2019short}, Moments in Time (MiT) \cite{monfort2018moments} and Human Action Clips Segments (HACS) Clips \cite{zhao2019hacs} datasets. For the transfer learning task, we evaluate on the smaller UCF-101 \cite{soomro2012ucf101} dataset.

\subsection{Training}
\label{sec:results:setup}

% Environment parameters
For HACS models are trained from random initialization (``from scratch'') without pre-training. We set the mini-batch size to 16 clips per GPU, with the total mini-batch size of 64. All experiments were performed with half-precision (float16) for more effective utilization of memory. Similar to relevant works \cite{feichtenhofer2019slowfast,feichtenhofer2020x3d}, we use a cosine-based learning rate decay schedule \cite{loshchilov2016sgdr} with the the learning rate $lr_{n}$ for iteration $n$ calculated as $lr_{n} = lr_{0} * 0.5[cos(\frac{n}{n_{max}}\pi)+1]$ in which $n_{max}$ is the total number of iterations and $lr_{0}$ is the starting learning rate. We use $lr_{0}=1.16$ and learning rate warm-up for the first 8k iterations similar to \cite{feichtenhofer2019slowfast,feichtenhofer2020x3d}. Batch sizes are determined by a multigrid method \cite{wu2020multigrid} with the initial batch size of 64. The multigrid learning rate follows the linear scaling rule \cite{goyal2017accurate} given the mini-batch size scaling. Unless specified otherwise, all experiments were performed over $n_{max} = 400$ epochs with momentum of 0.9 and weight decay of $5 \! \times \! 10^{-5}$.

For K-400, K-700 and MiT, we use similar training parameters but initialize the networks weights from the models trained on HACS. For the transfer learning task on UCF-101, we reduce the start learning rate to $0.01$ and do not use warm-up. We also decrease the number of epochs to 150 while including a learning rate multiplier for convolutional weights of value $0.1$. Unless stated otherwise, our models use $\delta=0.875$. We motivate this choice and experiment with other values in \Cref{sec:results:ablation}.

% Interval selection
The input frames are uniformly randomly selected. Based on the average clip length, for each dataset, we use equivalently sized temporal strides when selecting frames. For HACS with an average clip length of 60 frames, we use a temporal stride of 2. The two Kinetics datasets have clips with 250 frames on average and we use a temporal stride of 5. For MiT, we use a temporal stride of 3 with an average clips length of 90 frames. For UCF-101 we use strides of 4. On the spatial domain, we randomly crop a region of size $256 \! \times \! 256$ pixels in resized video frames with the shortest side being 320.

\subsection{Computational Inference}
\label{sec:results:inference}

% GFLOPS x views
We report inference with two different measures. We first report computational costs (FLOPs) similar to \cite{fan2019more,feichtenhofer2019slowfast,feichtenhofer2020x3d,tran2019video} by sampling 10 clips from a single video and perform 3 crops along the spatial dimensions ($10 \!\times \! 3 = 30$ views) of size $256 \! \times \! 256$. The inference time is then reported as the number of FLOPs per spatio-temporal view (clips times crops). This provides a standardized measure of computing inference when comparing across models as shown for \Cref{table:K400_accuracies,table:K700_accuracies,table:HACS_accuracies}.

\begin{table}[t]
\caption{\textbf{Comparison with K-400 state-of-the-art}. For consistency with previous testing methods, we report the model complexity as the GFLOPs per single clip view $\! \times \!$ the number of clips with spatial cropping of size $256 \! \times \! 256$.}
\centering
\resizebox{.5\textwidth}{!}{%
\renewcommand{\arraystretch}{1.5}
\begin{tabular}{c|c|c|c|c|c|c}
\hline
Model & 
Input &
Backbone &
top-1 (\%) & top-5 (\%) & 
GFLOPs $\! \times \!$ views &
Params\\
\hline

R(2+1)D \cite{tran2018closer} &
$16 \! \times \! 224^{2}$ &
ResNet101 &
62.8 & 83.9 &
$152 \! \times \! 115$&
63.6M\\

I3D \cite{carreira2017quo}  &
$16 \! \times \! 224^{2}$ &
InceptionV1 &
71.6 & 90.0 &
$108 \! \times \! N/A$&
12.0M\\

MF-Net \cite{chen2018multifiber} &
$16 \! \times \! 224^{2}$ &
ResNet50 &
72.8 & 90.4 &
$11.1 \! \times \! 50$ &
\textbf{8.0M}\\

TAM \cite{fan2019more} &
$(24,2) \! \times \! 224^{2}$ &
ResNet50 &
73.5 & 91.2 &
$ 93.4 \! \times \! 9$ &
25.0M\\

SRTG-101 (3D) \cite{stergiou2020learn} &
$16 \! \times \! 224^{2}$ &
ResNet101 &
73.2 & 91.3 &
$78.1 \! \times \! 30$ &
107.1M\\

SRTG-101 (2+1D) \cite{stergiou2020learn} &
$16 \! \times \! 224^{2}$ &
ResNet101 &
73.8 & 92.0 &
$163.1 \! \times \! 30$ &
105.3M\\

TSM \cite{lin2019tsm} &
$16 \! \times \! 224^{2}$ &
ResNet50 &
74.7 & 91.4 &
$65 \! \times \! 10$&
24.3M\\

ip-CSN-101 \cite{tran2019video} &
$8 \! \times \! 224^{2}$ &
ResNet101 &
76.7 & 92.3 &
$83 \! \times \! 30$ &
24.5M\\

ip-CSN-152 \cite{tran2019video} &
$8 \! \times \! 224^{2}$ &
ResNet152 &
77.8 & 92.8 &
$108.8 \! \times \! 30$ &
32.8M\\

SF-50 \cite{feichtenhofer2019slowfast} &
$(8,8) \! \times \! 224^{2}$ &
ResNet50 &
77.0 & 92.6 &
$65.7 \! \times \! 30$ &
34.4M\\

SF-101  \cite{feichtenhofer2019slowfast} &
$(8,8) \! \times \! 224^{2}$ &
ResNet101 &
77.9 & 93.5 &
$213 \! \times \! 30$ &
53.7M\\

SF-101+NL \cite{feichtenhofer2019slowfast} &
$(8,8) \! \times \! 224^{2}$ &
ResNet101 &
78.7 & 93.5 &
$116 \! \times \! 30$ &
59.9M\\

X3D-XL \cite{feichtenhofer2020x3d} &
$16 \! \times \! 312^{2}$ &
ResNet(X3D) &
\textbf{79.1} & \textbf{93.9} &
$48.4 \! \times \! 30$ &
11.0M\\
\hline

MTNet$_{S}$ \textbf{(ours)} &
$16 \! \times \! 256^{2}$ &
ResNet(X3D) &
74.8 & 92.1 &
$\mathbf{5.8 \! \times \! 30}$ &
25.8M\\

MTNet$_{M}$ \textbf{(ours)} &
$16 \! \times \! 256^{2}$ &
ResNet(X3D) &
76.6 & 92.5 &
$8.8 \! \times \! 30$ &
25.8M\\

MTNet$_{L}$ \textbf{(ours)} &
$16 \! \times \! 256^{2}$ &
ResNet(X3D) &
78.1 & 93.2 &
$17.6 \! \times \! 30$ &
50.1M\\

\end{tabular}%
}
\label{table:K400_accuracies}
\end{table}

% Forward and Backward time inference
In the case of architectural changes such as in \Cref{table:HACS_recurrent_cells} we additionally report the inference time in terms of computational latency (in msecs.) for forward and backward passes, independently. Computational latency times are calculated on single clips of size $16 \! \times \! 256 \! \times \! 256$.

\subsection{Main Results}
\label{sec:results:sota}

We discuss the comparisons of our MTNets to the current state-of-the-art for datasets K-400, MiT, K-700, and HACS.

% Kinetics-400
\textbf{Kinetics-400 (K-400)}. We present results in \Cref{table:K400_accuracies}. In comparison to the top performing X3D-XL \cite{feichtenhofer2020x3d}, our largest \textbf{MTNet$_{L}$} produces comparable performance (1.0\% top-1 and 0.7\% top-5 lower accuracies), despite a considerable reduction in computation of $\! \times  2.75$ in terms of GFLOPs. When comparing \textbf{MTNet$_{L}$}, we observe performance on par with the significantly larger SlowFast-101 \cite{feichtenhofer2019slowfast} which requires more than $\! \times  12$ the number of computations. Notably, \textbf{MTNet$_{L}$} outperforms the similarly complex MFNet \cite{chen2018multifiber} with +5.3\% top-1 and +2.8\% top-5 accuracies.

For the smaller \textbf{MTNet$_{M}$} we report accuracies close to \textit{Channel-Separated Network} (ip-CSN-101) \cite{tran2019video} and \textit{Temporal Adaptive Module} ResNet-50 \cite{fan2019more}, while being significantly more efficient than both. Considering its low number floating point operations (FLOPs), \textbf{MTNet$_{M}$} can still outperform R(2+1)D ResNet101 \cite{tran2018closer}, \textit{Temporal Shift Module} (TSM) \cite{lin2019tsm} and \textit{Squeeze and Recursion Temporal Gates} (SRTG) \cite{stergiou2020learn}. 

Finally, our smallest network \textbf{MTNet$_{S}$} performs on par with TSM and SRTG, while having the lowest number of FLOPs from all tested networks. %We note that the produced MTNets are the only family of spatio-temporal architectures within the range of sub-20 GFLOPs that produce accuracy rates similar to that of the top-performing and significantly more expensive state-of-the-art models. 

\begin{table}[t]
\caption{\textbf{Spatio-temporal block comparison on K-400}. Using a ResNet-50 as backbone, accuracy rates are reported for different spatio-temporal blocks. Numbers in parentheses are in comparison to 3D baseline.}
\centering
\resizebox{.5\textwidth}{!}{%
\renewcommand{\arraystretch}{1.5}
\begin{tabular}{c|ll|l|l}
\hline
Method & 
top-1 (\%) &
top-5 (\%) &
FLOPs (G) &
Params (M)\\
\hline

3D \cite{hara2018can} &
61.3  &
83.1  &
53.2  &
36.7  \\[0.25em]

(2+1)D \cite{tran2018closer} &
61.8 \textcolor{applegreen}{($+0.5$)} &
83.5 \textcolor{applegreen}{($+0.4$)} &
56.0 \textcolor{cadmiumred}{($+2.8$)} &
38.8 \textcolor{cadmiumred}{($+2.1$)} \\[0.25em]

Multi-Fiber \cite{chen2018multifiber} &
72.8 \textcolor{applegreen}{($+11.5$)} &
90.4 \textcolor{applegreen}{($+7.3$)} &
\textbf{22.5} \textcolor{applegreen}{($-30.7$)} &
\textbf{8.0} \textcolor{applegreen}{($-28.7$)} \\[0.25em]

Slow-only \cite{feichtenhofer2019slowfast} &
72.6 \textcolor{applegreen}{($+11.5$)} &
90.3 \textcolor{applegreen}{($+7.2$)} &
27.3 \textcolor{applegreen}{($-25.9$)} &
26.6 \textcolor{applegreen}{($-10.1$)} \\[0.25em]

SlowFast \cite{feichtenhofer2019slowfast} &
74.3 \textcolor{applegreen}{($+13$)} &
91.0 \textcolor{applegreen}{($+7.9$)} &
39.8 \textcolor{applegreen}{($-13.4$)} &
34.4 \textcolor{applegreen}{($-2.3$)} \\[0.25em]

\hline

MTConv \textbf{(ours)} &
\textbf{74.8} \textcolor{applegreen}{($+13.5$)} &
\textbf{91.3} \textcolor{applegreen}{($+8.2$)} &
23.1 \textcolor{applegreen}{($-30.1$)} &
35.7 \textcolor{applegreen}{($-1.0$)} \\[0.25em]

\end{tabular}%
}
\label{table:K400_accuracies_Res50}
\end{table}

\begin{table}[t]
\caption{\textbf{Comparison with MiT state-of-the-art}. Models denoted with $^{\ddagger}$ include additional optical flow input.}
\centering
\resizebox{.7\linewidth}{!}{%
\renewcommand{\arraystretch}{1.5}
\begin{tabular}{c|c|cc}
\hline
Model &
Arch. size &
top-1 (\%) & top-5 (\%) \\
\hline

EvaNet \cite{piergiovanni2019evolving} &
\multirow{2}{*}{NAS\cite{zoph2017neural}} &
31.8 &	N/A \\

AssembleNet \cite{ryoo2019assemblenet} & &
34.3 &	\textbf{62.7}\\

\hline

TSN-Flow \cite{monfort2018moments} $^\ddagger$  &
\multirow{8}{*}{Fixed}&
15.7 & 34.7 \\

TSN-2stream \cite{monfort2018moments} $^\ddagger$ & &
25.3 & 50.1\\

TRN-Multiscale \cite{zhou2018temporal} & &
28.3 & 53.9\\

I3D \cite{carreira2017quo} & &
29.5 &	56.1 \\

CoST \cite{li2019collaborative} & &
32.4 & 60.0\\

SRTG-101 \cite{stergiou2020learn} & &
33.6 & 58.5 \\\cline{1-1}\cline{3-4}

MTNet$_{M}$ \textbf{(ours)} & &
34.5 & 58.6 \\

MTNet$_{L}$ \textbf{(ours)} & &
\textbf{35.2} & 59.3 \\

\end{tabular}%
}
\label{table:mit_accuracies}
\end{table}

% Block changes on ResNet 50 on Kinetics
To better understand the relative contribution of network architecture and convolution operator, we compare a variety of spatio-temporal convolutional methods on the same ResNet-50 architecture. As shown in \Cref{table:K400_accuracies_Res50}, a direct replacement to MTConvs can yield a significant performance improvement over 3D convolutions with +13.5\% in top-1 and +8.2\% in top-5 accuracies. MTConvs also outperform other popular spatio-temporal convolution-based methods. MTConvs reduce the number of GFLOPs by 56\% in comparison to regular 3D convolutions. The decrease in FLOPs for MTConvs does not come at a cost of parameters. MTNets include only a slightly reduced number of parameters in comparison to 3D Convs, which allows the models to preserve the level of complexity.

% MiT results
\textbf{Moments in Time (MiT)}. \Cref{table:mit_accuracies} summarizes performance in terms of the top-1 and top-5 accuracies of current state-of-the-art models. Comparisons are performed on models with fixed-sized architectures as well as those that employ Neural Architecture Search (NAS) \cite{zoph2017neural}. Our best performing architecture \textbf{MTNet$_{L}$} outperforms current state-of-the-art models with top-1 accuracy of 35.2\%. Notably, these comparisons also include models with supplementary inputs optical flow \cite{monfort2018moments} and audio \cite{monfort2018moments} while both MTNet architectures are trained only on RGB frames. The smaller \textbf{MTNet$_{M}$} achieves a similar classification accuracy compared to learned architectures such as AssembleNet \cite{piergiovanni2019evolving}. This comes with a reduction in terms of computations as there is no additional objective to permute the base model.

\begin{table}[t]
\caption{\textbf{Comparison with K-700 state-of-the-art}. GFLOP calculation is similar to that in \Cref{table:K400_accuracies}.}
\centering
\resizebox{.5\textwidth}{!}{%
\renewcommand{\arraystretch}{1.5}
\begin{tabular}{c|c|cc|c}
\hline
Model & 
Pre-train &
top-1 (\%) & 
top-5 (\%) &
GFLOPs $\! \times \!$ views \\
\hline

I3D \cite{carreira2017quo} &
K-600 &
58.7 & 81.7 &
$108 \! \times \! N/A$ \\

SRTG-101 (3D) \cite{stergiou2020learn} &
HACS &
56.5 & 76.8  &
$78.1 \! \times \! 30$ \\

SRTG-101 (2+1)D \cite{stergiou2020learn} &
HACS &
56.8 & 77.4 &
$163.1 \! \times \! 30$ \\
\hline

MTNet$_{M}$ \textbf{(ours)} &
HACS &
58.4 & 77.6 &
$8.8 \! \times \! 30$ \\

MTNet$_{L}$ \textbf{(ours)} &
HACS &
\textbf{63.3} & \textbf{84.1} &
$17.6 \! \times \! 30$ \\

\end{tabular}%
}
\label{table:K700_accuracies}
\end{table}

\begin{table}[t]
\caption{\textbf{Comparison  with  HACS  state-of-the-art}. Weight initialization sources are denoted by their respective indicators.}
\begin{threeparttable}[t]
\centering
\resizebox{.5\textwidth}{!}{%
\renewcommand{\arraystretch}{1.2} 
\begin{tabular}{c|c|c|c|c|c}
\hline
Model & 
Pre-train &
top-1 & top-5 & 
GFLOPs $\! \times \!$ views &
Params \\
\hline

MF-Net \cite{chen2018multifiber}$^{\dagger}$ &
\multirow{4}{*}{K-400} &
78.3 & 94.6 &
$11.1 \! \times \! 50$ &
8.0M \\

TAM \cite{fan2019more}$^{\dagger}$ &
 &
82.2 & 95.2 &
$93.4 \! \times \! 9$ &
25.0M\\

SF-101 \cite{feichtenhofer2019slowfast}$^{\dagger}$ &
 &
83.7 & 96.8 &
$213 \! \times \! 30$ &
53.7M \\

X3D-L \cite{feichtenhofer2020x3d}$^{\dagger}$ &
&
85.8 & 96.1 &
$24.8 \! \times \! 30$ &
6.1M\\
\hline

R3D-101 \cite{kataoka2020would}$^{*}$ &
\multirow{2}{*}{K-700} &
80.5 & 95.8 &
$78.0 \! \times \! 30$ &
69.0M\\

R(2+1)D-101 \cite{kataoka2020would}$^{\dagger}$ &
 &
82.9 & 95.6 &
$163.0 \! \times \! 30$ &
72.1M\\
\hline

ir-CSN-101 \cite{tran2019video}$^{\dagger}$ &
\multirow{2}{*}{IG65} &
83.8 & 93.8 &
$63.6 \! \times \! 10$ &
22.1M \\

ip-CSN-101 \cite{tran2019video}$^{\dagger}$ &
 &
84.1 & 93.9 &
$63.6 \! \times \! 10$ &
24.5M \\
\hline

SRTG-101 (3D) \cite{stergiou2020learn}$^{\dagger}$ &
- &
81.6 & 96.3 &
$78.1 \! \times \! 30$ &
107.1M \\

SRTG-101 (2+1)D \cite{stergiou2020learn}$^{\dagger}$ &
- &
84.3 & \textbf{96.8} &
$163.1 \! \times \! 30$ &
105.3M \\
\hline

MTNet$_{S}$ \textbf{(ours)} &
- &
80.7 & 95.2 &
$5.8 \! \times \! 30$  &
25.8M\\

MTNet$_{M}$ \textbf{(ours)} &
- &
83.4 & 95.9 &
$8.8 \! \times \! 30$  &
25.8M\\

MTNet$_{L}$ \textbf{(ours)} &
- &
\textbf{86.6} & \textbf{96.7} &
$17.6 \! \times \! 30$  &
50.1M\\

\end{tabular}%
}
 \begin{tablenotes}
    \item[$\dagger$] models and weights from authors' repositories.  
    \item[$*$] models and weights that we re-trained. 
   \end{tablenotes}
\end{threeparttable}%
\label{table:HACS_accuracies}
\end{table}

\begin{table*}%[H]
           \centering
           \captionsetup[subtable]{position = top}
           \caption{\textbf{Ablation studies on HACS.} We evaluate MTNet architectures under different training parameters and report top-1 and top-5 accuracies as well as the number of GFLOPs and parameters.}
           \begin{subtable}{0.5\linewidth}
           \caption{\textbf{Branch channel ratio}: Varying channel ratio ($\delta$) across MTNet$_{M}$ and MTNet$_{L}$ architectures.}
               \centering
               \resizebox{.85\linewidth}{!}{%
               \begin{tabular}{p{0.3cm}| l |cc|c|c}
                \hline
                    Net. &
                    $\delta$ setting &
                    top-1 &
                    top-5 &
                    GFLOPs &
                    Params (M)\\ [0.15ex]
                    \hline
                    \multirow{7}{*}{\rotatebox[origin=c]{90}{MTNet$_{M}$}} &  $\delta=1.0$ (No $\mathcal{P}$) & 82.2 & 93.6 & 10.8 & 29.7\\
                                             [0.5ex] 
                                             & $\qquad 7/8$ & \textbf{83.4} & \textbf{95.9} & 8.8 & 25.8\\ 
                                             [0.5ex]
                                             & $\qquad 3/4$ & 83.1 & 95.6 & 6.7 & 21.8\\ 
                                             [0.5ex]
                                             & $\qquad 5/8$ & 81.6 & 93.2 & 4.8 & 19.3\\ 
                                             [0.5ex]
                                             & $\qquad 1/2$ & 79.7 & 91.8 & 3.6 & \textbf{18.6}\\ 
                                             [0.5ex]
                                             & $\qquad 3/8$ & 78.6 & 89.4 & 2.6 & 19.2\\ 
                                             [0.5ex]
                                             & $\qquad 1/4$ & 77.1 & 88.6 & \textbf{2.1} & 21.0\\ 
                                             [0.5ex]
                    \hline
                    \multirow{6}{*}{\rotatebox[origin=c]{90}{MTNet$_{L}$}} &  $\delta=1.0$ (No $\mathcal{P}$) & 84.9 & 95.7 & 20.6 & 53.5\\
                                             [0.5ex] 
                                             & $\qquad 7/8$ & \textbf{86.6} & \textbf{96.7} & 17.6 & 50.1\\ 
                                             [0.5ex]
                                             & $\qquad 3/4$ & 86.1 & 96.2 & 12.5 & 45.3\\ 
                                             [0.5ex]
                                             & $\qquad 1/2$ & 83.2 & 95.3  & 7.09 & \textbf{42.7}\\ 
                                             [0.5ex]
                                             & $\qquad 3/8$ & 82.1 & 93.9 & 5.2 & 45.3\\ 
                                             [0.5ex]
                                             & $\qquad 1/4$ & 80.3 & 92.4 & \textbf{4.1} & 47.8\\ 
                                             [0.5ex]
                \end{tabular}%
                }
               \label{table:HACS_delta}
           \end{subtable}%
           \hspace*{1em}
           \begin{subtable}{0.5\linewidth}
                \vspace*{-10em}
               \caption{\textbf{Recurrent cell configurations:} Alternative recurrent cells for the \textit{global aggregated feature importance branch} ($\mathcal{G}$). Branch ratio of $\delta=7/8$ is used unless otherwise stated.}
               \centering
               \resizebox{\linewidth}{!}{%
               \begin{tabular}{c|c|cc|cc|cc}
                \hline
                \multirow{2}{*}{Net} &
                \multirow{2}{*}{Cell type} & 
                Params &
                FLOPS &
                \multicolumn{2}{c|}{Latency (msec)} &
                \multirow{2}{*}{top-1} & 
                \multirow{2}{*}{top-5}\\
                &
                & 
                (M) &
                (G) &
                $\downarrow$F & $\uparrow$B &
                &
                \\[0.3em]
                \hline
                \parbox[t]{2mm}{\multirow{5}{*}{\rotatebox[origin=c]{90}{MTNet$_S$}}} &
                RNN \cite{rumelhart1985learning} &
                24.3 &
                5.8 &
                58 & 78 &
                78.8 &
                93.7\\ [0.25em]
                &
                LSTM \cite{hochreiter1997long} &
                26.5 &
                5.8 &
                61 & 79 &
                79.9 &
                94.3\\[0.25em]
                &
                LSTM (peephole) \cite{gers2000recurrent} &
                26.5 &
                5.8 &
                68 & 85 &
                80.1 &
                94.5\\[0.25em]
                &
                GRU \cite{cho2014learning} &
                25.8 &
                5.8 &
                65 & 80 &
                \textbf{80.7} &
                \textbf{95.2}\\[0.25em]
                \hline
                \parbox[t]{2mm}{\multirow{5}{*}{\rotatebox[origin=c]{90}{MTNet$_M$}}} &
                RNN \cite{rumelhart1985learning} &
                24.3 &
                8.8 &
                84 & 113 &
                82.5 &
                94.8\\ [0.25em]
                &
                LSTM \cite{hochreiter1997long} &
                26.5 &
                8.8 &
                86 & 109 &
                83.1 &
                95.4\\[0.25em]
                &
                LSTM (peephole) \cite{gers2000recurrent} &
                26.5 &
                8.8 &
                94 & 120 &
                83.2 &
                95.6\\[0.25em]
                &
                GRU \cite{cho2014learning} &
                25.8 &
                8.8 &
                90 & 111 &
                \textbf{83.4} & 
                \textbf{95.9} \\[0.25em]
                \end{tabular}%
                }
                 \label{table:HACS_recurrent_cells}
           \end{subtable}%
           \hspace*{-33em}
           \begin{subtable}{0.5\linewidth}
                \vspace*{18em}
                    \caption{\textbf{Spatio-temporal pooling methodology:} Top-1 accuracy for different pooling methods used on inputs for the \textit{prolonged branch} ($\mathcal{P}$).}
                    \centering
                    \resizebox{\linewidth}{!}{%
                    \begin{tabular}{c|cccccc}
                        \hline
                        \multirow{2}{*}{Net} &
                        \multicolumn{6}{c}{Pooling} \\[0.2em]\cline{2-7}
                        &
                        Avg &
                        Max &
                        Stochastic \cite{zeiler2013stochastic} &
                        SoftPool \cite{stergiou2021refining} &
                        Avg + $cos$ &
                        SoftPool \cite{stergiou2021refining} + $cos$ \\
                        \hline
                        MTNet$_{S}$ &
                        77.8 &
                        75.9 &
                        76.8 &
                        77.8 &
                        80.5 &
                        \textbf{80.7} \\[0.25em]
                        MTNet$_{M}$ &
                        79.8 &
                        77.6 &
                        78.2 &
                        80.7 &
                        82.6 &
                        \textbf{83.4} \\[0.25em]
                        MTNet$_{L}$ &
                        83.8 &
                        82.1 &
                        82.9 &
                        84.2 &
                        85.9 &
                        \textbf{86.6} \\[0.25em]
                    \end{tabular}%
                    }
                 \label{table:HACS_poling}
           \end{subtable}
\end{table*}

% Results on K-700
\textbf{Kinetics-700 (K-700)}. We further evaluate our MTNets and their generalization capabilities on the recently introduced 700-class variant of Kinetics. As shown in \Cref{table:K700_accuracies}, our architectures demonstrate similar performance trends as the accuracy values reported in \Cref{table:K400_accuracies,table:mit_accuracies}. Specifically, we observe that \textbf{MTNet$_{L}$} outperforms other methods by a significant margin of +(4.7-6.8)\% for top-1 accuracy and +(2.4-7.3)\% for top-5. Again, \textbf{MTNet$_{M}$} performs similar to I3D \cite{carreira2017quo} with a strongly reduced number of GFLOPs.

\textbf{HACS}. Finally, we present results on HACS in \Cref{table:HACS_accuracies}. The datasets on which the models have been pre-trained are included in the table. Note that MTNets are trained on HACS from scratch. As shown, the use of MTConvs improves the overall accuracy. Notably, \textbf{MTNet$_{S}$} performs similarly to both R3D-101 \cite{kataoka2020would} and SRTG-101 (3D) \cite{stergiou2020learn}. \textbf{MTNet$_{M}$} provides overall higher performance with additional +2.7\% and +0.7\% top-1 and top-5 accuracies over the smaller counterpart \textbf{MTNet$_{S}$}, while achieving similar accuracies as SlowFast-101 and ir-CSN-101. Finally, comparing \textbf{MTNet$_{L}$} to X3D-L shows an improvement of +0.8\% for the top-1 and +0.6\% top-5 accuracies while having $\sim\!$ 29\% less FLOPs.

\subsection{Ablation Studies}
\label{sec:results:ablation}

In this section we provide ablation studies on the HACS dataset. We compare different ratios ($\delta$) used by the local ($\mathcal{L}$) and prolonged ($\mathcal{P}$) branches. We additionally evaluate the effect of different recurrent cells on the global aggregated feature importance branch ($\mathcal{G}$). Finally, we present results based on different spatio-temporal pooling methods applied to inputs of $\mathcal{P}$. 

% experiments with ratio
\textbf{Branch channel ratio}. As shown in \Cref{table:HACS_delta}, the best performing ratios ($\delta$) are within the range of (0.875 $\! \sim \!$ 0.75) with marginal differences in the range of $\pm (0.3 \! \sim \! 0.5)\%$ for both the top-1 and top-5 performances. These ratios also lead to a reduction in computational costs and the number of parameters. Improvements on number of computations (GFLOPs) based on these ratios are shown by the reduction of (25 $\! \sim \!$ 37)\% when using both $\mathcal{L}$ and $\mathcal{P}$ branches, compared to using solely the local branch ($\mathcal{L}$) which is equivalent to a single standard 3D Conv. We attribute the loss in performance when using small ratios to the dependency of branch $\mathcal{P}$ on branch $\mathcal{L}$. Interestingly, the decrease in feature dimensionality of the local features with the use of smaller $\delta$ values corresponds to the inability of the prolonged features to encapsulate substantial video action details by themselves. In addition, decreases in $\delta$ do not directly relate to decreases in the number of parameters as seen in \Cref{table:HACS_delta}. For $\delta < 1/2$, the number of parameters increases again with branch $\mathcal{P}$ employing a larger number of parameters. Therefore, the smallest number of parameters is observed when the ratio is split equally between the two channels ($\delta=1/2$). This setting shows the largest combined reduction of GFLOPs (-66\%) and number of parameters (-37\%) for a standard 3D Conv. Lastly, we note that zero ratios $\delta=0$ are not feasible as branch $\mathcal{P}$ includes the outputs of branch $\mathcal{L}$ which thus cannot be omitted.  

% recurrent cells
\textbf{Recurrent cell configuration}. Next, we study the effect that the recurrent cell methodology used in $\mathcal{G}$ has on the accuracy. Recurrent layers are replaced in \textbf{MTNet$_{S}$} and \textbf{MTNet$_{M}$} with the changes only affecting branch $\mathcal{G}$. Latencies are calculated as the time (in msecs.) required for a full forward ($\downarrow$F) and backward ($\uparrow$B) pass for a single clip of size $16 \! \times \! 256 \! \times \! 256$. Results appear in \Cref{table:HACS_recurrent_cells}. The proposed use of GRUs \cite{cho2014learning} is motivated by the (slight) improvements over alternative recurrent cell structures. For \textbf{MTNet$_{S}$}, GRUs perform better than regular RNN cells \cite{rumelhart1985learning} with +1.9\% top-1 and +1.5\% top-5 accuracies. However, the overall simplicity of RNNs can be more efficient in terms of parameter use with a -8\% overall network parameter reduction as well as marginally faster forward and backward latency times. A similar observation is made for \textbf{MTNet$_{M}$} as GRU's top-1 and top-5 accuracies improve the RNN baseline by +0.9\% and +1.1\% respectively. Compared to LSTMs \cite{hochreiter1997long} and LSTMs with peepholes variants \cite{gers2000recurrent}, GRUs also show marginally better accuracy rates. We note that a property of GRUs is the merge of LSTM's \textit{forget} and \textit{input} states as well as their \textit{cell} and \textit{hidden} states. This simplifies the recurrent structure.

% Table: UCF accuracies
\begin{table}[t]
\caption{\textbf{Transfer learning on UCF-101:} Top-1 and top-5 accuracies after pre-training.}
\centering
\resizebox{.45\textwidth}{!}{%
\begin{tabular}{c|c|cc}
\hline
Model & 
Pre-training &
top-1 (\%) & top-5 (\%) \\[0.3em]
\hline
I3D & 
K-400 &
92.4 & 97.6 \\[0.3em]

TSM &
K-400 &
92.3 & 97.9 \\[0.3em]

ir-CSN-152 &
IG65M &
95.4 & 99.2 \\[0.3em]

MF-Net &
K-400 &
93.8 & 98.4 \\[0.3em]

SF-50 &
ImageNet &
94.6 & 98.7 \\[0.3em]

SF-101 &
ImageNet &
95.8 & 99.1 \\[0.3em]

SRTG-101 (2+1)D &
HACS+K-700 &
97.2 & 99.1 \\[0.3em]

SRTG-101 (3D) &
HACS+K-700 &
97.3 & \textbf{99.6} \\[0.3em]

\hline
MTNet$_{S}$ \textbf{(ours)} &
HACS &
94.2 & 98.0 \\[0.3em]

MTNet$_{M}$ \textbf{(ours)} &
HACS &
95.4 & 98.1 \\[0.3em]

MTNet$_{L}$ \textbf{(ours)} &
HACS &
\textbf{97.4} & 99.2 \\
\end{tabular}%
}
\label{table:accuracies_ucf}
\end{table}

% Pooling methods
\textbf{Spatio-temporal pooling methodology}. We conclude our ablation studies by exploring the effect of different pooling methods used for $\mathcal{P}$ branch's inputs. Experiments were performed with temporal and spatial symmetric and asymmetric methods. In the first category, operations are performed similarly in all dimensions while the latter methods perform spatial and temporal pooling independently. In \Cref{table:HACS_poling}, we report the top-1 accuracies for different pooling configurations. Frame selection with the proposed temporal triplet cosine ($cos$) similarity yields overall improvements over symmetric methods. For average pooling with triplet $cos$, accuracies are improved by +2.7\% for \textbf{$MTNet_{S}$}, +2.8\% for \textbf{$MTNet_{M}$} and +2.1\% for \textbf{$MTNet_{L}$}. Similarly, using SoftPool\cite{stergiou2021refining} and triplet $cos$ increases top-1 accuracy, by +2.9\%, +2.7\% and +2.4\% for each of the models respectively, in comparison to symmetric SoftPool. The improvement of SoftPool asymmetrically compared to average pooling is only marginal with +0.57\% improvement on accuracy on average across the three architectures. We thus conclude that the temporal dimensionality reduction method has a significantly larger effect on the overall performance than the selection of a spatial method. Temporal reductions with asymmetric methods such as our combined SoftPool with triplet cosine similarity shows greater accuracy gains than symmetric methods that apply a pooling operation across all dimensions.

\subsection{Feature Transferability with MTNets}
\label{sec:results:TL}

% Transfer learning set-up
We compare the transfer learning capabilities of MTNets with state-of-the-art video models on the smaller action recognition dataset UCF-101. To allow a fair comparison with other methods, all tested architectures are initialized with weights as in  \Cref{table:HACS_accuracies}. \textbf{$MTNet_{L}$} achieves performance comparable to that of SRTG-101 (2+1)D which had been pre-trained on both large-scale HACS and K-700 datasets. The second model \textbf{$MTNet_{M}$} can also perform as well as the ir-CSN that used a 65M dataset sourced from Instagram \cite{ghadiyaram2019large} and the 101-layer variant of SlowFast. Our smallest architecture \textbf{$MTNet_{S}$} also shows good performance with accuracies above those of TSM and I3D while being similar to SlowFast-50. This further shows the generalization capabilities of our varying spatio-temporal feature extraction approach.

\section{Conclusions}
\label{sec:conclusions}

% General remarks
We have introduced a novel multi-temporal convolution (MTConv) block that models variations in the performance in action videos by extracting and aligning spatio-temporal patterns across temporal scales. Our proposed convolution block uses two branches to address motions performed within a short and prolonged time span, respectively. A third global aggregated feature importance branch aligns the output activations of the first two branches based on the discovered feature dynamics. With this mechanism, we can extract salient spatio-temporal patterns despite potential differences in the temporal execution. We have also introduced MTNets that include MTConvs in a X3D backbone. MTNets achieve comparable or, in many cases, higher classification accuracies than current state-of-the-art models on the most widely used action recognition benchmarks. Importantly, MTNets achieve a reduction in terms of computation costs. Based on these results, we believe that the modeling of variable-duration spatio-temporal patterns can be more widely exploited in future research in the field of video action recognition.

\section{Acknowledgments}
\label{sec:acknowledgments}
This publication is supported by the Netherlands Organization for Scientific Research (NWO) with a TOP-C2 grant for Automatic recognition of bodily interactions (ARBITER).
% TODO RONALD: leave out for now

{\small
\bibliographystyle{IEEEtran}
\bibliography{egbib}

\begin{thebibliography}{10}
\providecommand{\url}[1]{#1}
\csname url@samestyle\endcsname
\providecommand{\newblock}{\relax}
\providecommand{\bibinfo}[2]{#2}
\providecommand{\BIBentrySTDinterwordspacing}{\spaceskip=0pt\relax}
\providecommand{\BIBentryALTinterwordstretchfactor}{4}
\providecommand{\BIBentryALTinterwordspacing}{\spaceskip=\fontdimen2\font plus
\BIBentryALTinterwordstretchfactor\fontdimen3\font minus
  \fontdimen4\font\relax}
\providecommand{\BIBforeignlanguage}[2]{{%
\expandafter\ifx\csname l@#1\endcsname\relax
\typeout{** WARNING: IEEEtran.bst: No hyphenation pattern has been}%
\typeout{** loaded for the language `#1'. Using the pattern for}%
\typeout{** the default language instead.}%
\else
\language=\csname l@#1\endcsname
\fi
#2}}
\providecommand{\BIBdecl}{\relax}
\BIBdecl

\bibitem{stergiou2019analyzing}
A.~Stergiou and R.~Poppe, ``Analyzing human-human interactions: A survey,''
  \emph{Computer Vision and Image Understanding}, vol. 188, p. 102799, 2019.

\bibitem{dong1995statistics}
D.~W. Dong and J.~J. Atick, ``Statistics of natural time-varying images,''
  \emph{Network: Computation in Neural Systems}, vol.~6, no.~3, pp. 345--358,
  1995.

\bibitem{ji20133d}
S.~Ji, W.~Xu, M.~Yang, and K.~Yu, ``{3D} convolutional neural networks for
  human action recognition,'' \emph{Transactions on Pattern Analysis and
  Machine Intelligence}, vol.~35, no.~1, pp. 221--231, 2013.

\bibitem{vallacher2011action}
R.~R. Vallacher and D.~M. Wegner, ``Action identification theory,''
  \emph{Handbook of theories of social psychology}, vol.~1, pp. 327--349, 2011.

\bibitem{carreira2017quo}
J.~Carreira and A.~Zisserman, ``Quo vadis, action recognition? {A} new model
  and the {K}inetics dataset,'' in \emph{Computer Vision and Pattern
  Recognition (CVPR)}, 2017, pp. 4724--4733.

\bibitem{monfort2018moments}
M.~Monfort, A.~Andonian, B.~Zhou, K.~Ramakrishnan, S.~A. Bargal, T.~Yan,
  L.~Brown, Q.~Fan, D.~Gutfreund, C.~Vondrick, and A.~Oliva, ``Moments in time
  dataset: {O}ne million videos for event understanding,'' \emph{IEEE
  Transactions on Pattern Analysis and Machine Intelligence}, vol.~42, no.~2,
  pp. 502--508, 2019.

\bibitem{carreira2019short}
J.~Carreira, E.~Noland, C.~Hillier, and A.~Zisserman, ``A short note on the
  {Kinetics-700} human action dataset,'' \emph{arXiv preprint
  arXiv:1907.06987}, 2019.

\bibitem{zhao2019hacs}
H.~Zhao, A.~Torralba, L.~Torresani, and Z.~Yan, ``{HACS}: Human action clips
  and segments dataset for recognition and temporal localization,'' in
  \emph{International Conference on Computer Vision (ICCV)}, 2019, pp.
  8668--8678.

\bibitem{soomro2012ucf101}
K.~Soomro, A.~R. Zamir, and M.~Shah, ``{UCF101}: A dataset of 101 human actions
  classes from videos in the wild,'' \emph{arXiv preprint arXiv:1212.0402},
  2012.

\bibitem{simonyan2014two}
K.~Simonyan and A.~Zisserman, ``Two-stream convolutional networks for action
  recognition in videos,'' in \emph{Advances in Neural Information Processing
  Systems (NIPS)}, 2014, pp. 568--576.

\bibitem{feichtenhofer2016spatiotemporal}
C.~Feichtenhofer, A.~Pinz, and R.~Wildes, ``Spatiotemporal residual networks
  for video action recognition,'' in \emph{Advances in Neural Information
  Processing Systems (NIPS)}, 2016, pp. 3468--3476.

\bibitem{wang2016temporal}
L.~Wang, Y.~Xiong, Z.~Wang, Y.~Qiao, D.~Lin, X.~Tang, and L.~Van~Gool,
  ``Temporal segment networks: Towards good practices for deep action
  recognition,'' in \emph{European Conference on Computer Vision (ECCV)}, 2016,
  pp. 20--36.

\bibitem{diba2017deep}
A.~Diba, V.~Sharma, and L.~Van~Gool, ``Deep temporal linear encoding
  networks,'' in \emph{Computer Vision and Pattern Recognition (CVPR)}, 2017,
  pp. 2329--2338.

\bibitem{baccouche2011sequential}
M.~Baccouche, F.~Mamalet, C.~Wolf, C.~Garcia, and A.~Baskurt, ``Sequential deep
  learning for human action recognition,'' in \emph{International Workshop on
  Human Behavior Understanding (HBU)}, 2011, pp. 29--39.

\bibitem{hara2018can}
K.~Hara, H.~Kataoka, and Y.~Satoh, ``Can spatiotemporal {3D CNNs} retrace the
  history of {2D CNNs} and {I}mage{N}et?'' in \emph{Computer Vision and Pattern
  Recognition (CVPR)}, 2018, pp. 18--22.

\bibitem{kataoka2020would}
H.~Kataoka, T.~Wakamiya, K.~Hara, and Y.~Satoh, ``Would mega-scale datasets
  further enhance spatiotemporal {3D CNNs}?'' \emph{arXiv preprint
  arXiv:2004.04968}, 2020.

\bibitem{tran2018closer}
D.~Tran, H.~Wang, L.~Torresani, J.~Ray, Y.~LeCun, and M.~Paluri, ``A closer
  look at spatiotemporal convolutions for action recognition,'' in
  \emph{Conference on Computer Vision and Pattern Recognition (CVPR)}, 2018,
  pp. 6450--6459.

\bibitem{chen2018multifiber}
Y.~Chen, Y.~Kalantidis, J.~Li, S.~Yan, and J.~Feng, ``Multi-fiber networks for
  video recognition,'' in \emph{European Conference on Computer Vision (ECCV)},
  2018, pp. 352--367.

\bibitem{tran2019video}
D.~Tran, H.~Wang, L.~Torresani, and M.~Feiszli, ``Video classification with
  channel-separated convolutional networks,'' in \emph{International Conference
  on Computer Vision (ICCV)}.\hskip 1em plus 0.5em minus 0.4em\relax IEEE,
  2019, pp. 5552--5561.

\bibitem{feichtenhofer2020x3d}
C.~Feichtenhofer, ``X3d: Expanding architectures for efficient video
  recognition,'' in \emph{Conference on Computer Vision and Pattern Recognition
  (CVPR)}, 2020, pp. 203--213.

\bibitem{lin2019tsm}
J.~Lin, C.~Gan, and S.~Han, ``{TSM}: Temporal shift module for efficient video
  understanding,'' in \emph{International Conference on Computer Vision
  (ICCV)}, 2019, pp. 7083--7093.

\bibitem{sudhakaran2020gate}
S.~Sudhakaran, S.~Escalera, and O.~Lanz, ``Gate-shift networks for video action
  recognition,'' in \emph{Conference on Computer Vision and Pattern Recognition
  (CVPR)}, 2020, pp. 1102--1111.

\bibitem{feichtenhofer2019slowfast}
C.~Feichtenhofer, H.~Fan, J.~Malik, and K.~He, ``{SlowFast} networks for video
  recognition,'' in \emph{International Conference on Computer Vision (ICCV)},
  2019, pp. 6202--6211.

\bibitem{qiu2019learning}
Z.~Qiu, T.~Yao, C.-W. Ngo, X.~Tian, and T.~Mei, ``Learning spatio-temporal
  representation with local and global diffusion,'' in \emph{Conference on
  Computer Vision and Pattern Recognition (CVPR)}, 2019, pp. 12\,056--12\,065.

\bibitem{chen2019drop}
Y.~Chen, H.~Fan, B.~Xu, Z.~Yan, Y.~Kalantidis, M.~Rohrbach, S.~Yan, and
  J.~Feng, ``Drop an octave: Reducing spatial redundancy in convolutional
  neural networks with octave convolution,'' in \emph{International Conference
  on Computer Vision (ICCV)}, 2019, pp. 3435--3444.

\bibitem{hu2018squeeze}
J.~Hu, L.~Shen, and G.~Sun, ``Squeeze-and-excitation networks,'' in
  \emph{Conference on Computer Vision and Pattern Recognition (CVPR)}, 2018,
  pp. 7132--7141.

\bibitem{hu2018gather}
J.~Hu, L.~Shen, S.~Albanie, G.~Sun, and A.~Vedaldi, ``Gather-excite: Exploiting
  feature context in convolutional neural networks,'' in \emph{Advances in
  Neural Information Processing Systems (NeurIPS)}, 2018, pp. 9401--9411.

\bibitem{long2018attention}
X.~Long, C.~Gan, G.~De~Melo, J.~Wu, X.~Liu, and S.~Wen, ``Attention clusters:
  Purely attention based local feature integration for video classification,''
  in \emph{Conference on Computer Vision and Pattern Recognition (CVPR)}, 2018,
  pp. 7834--7843.

\bibitem{wang2018appearance}
L.~Wang, W.~Li, W.~Li, and L.~Van~Gool, ``Appearance-and-relation networks for
  video classification,'' in \emph{Conference on Computer Vision and Pattern
  Recognition (CVPR)}, 2018, pp. 1430--1439.

\bibitem{stergiou2020learn}
A.~Stergiou and R.~Poppe, ``Learn to cycle: Time-consistent feature discovery
  for action recognition,'' \emph{Pattern Recognition Letters}, vol. 141, pp.
  1--7, 2021.

\bibitem{liu2021tam}
Z.~Liu, L.~Wang, W.~Wu, C.~Qian, and T.~Lu, ``{TAM}: Temporal adaptive module
  for video recognition,'' \emph{arXiv preprint arXiv:2005.06803}, 2020.

\bibitem{ioffe2015batch}
S.~Ioffe and C.~Szegedy, ``Batch normalization: Accelerating deep network
  training by reducing internal covariate shift,'' in \emph{International
  Conference on Machine Learning (ICML)}, 2015, pp. 448--456.

\bibitem{stergiou2021refining}
A.~Stergiou, R.~Poppe, and K.~Grigorios, ``Refining activation downsampling
  with {SoftPool},'' \emph{arXiv preprint}, 2021.

\bibitem{cho2014learning}
K.~Cho, B.~van Merri{\"e}nboer, C.~Gulcehre, D.~Bahdanau, F.~Bougares,
  H.~Schwenk, and Y.~Bengio, ``Learning phrase representations using {RNN}
  encoder--decoder for statistical machine translation,'' in \emph{Conference
  on Empirical Methods in Natural Language Processing (EMNLP)}, 2014, pp.
  1724--1734.

\bibitem{radosavovic2020designing}
I.~Radosavovic, R.~P. Kosaraju, R.~Girshick, K.~He, and P.~Doll{\'a}r,
  ``Designing network design spaces,'' in \emph{Conference on Computer Vision
  and Pattern Recognition (CVPR)}, 2020, pp. 10\,428--10\,436.

\bibitem{loshchilov2016sgdr}
I.~Loshchilov and F.~Hutter, ``{SGDR:} {S}tochastic gradient descent with warm
  restarts,'' \emph{International Conference on Learning Representations
  (ICLR)}, 2017.

\bibitem{wu2020multigrid}
C.-Y. Wu, R.~Girshick, K.~He, C.~Feichtenhofer, and P.~Kr{\"a}henb{\"u}hl, ``A
  multigrid method for efficiently training video models,'' in \emph{Conference
  on Computer Vision and Pattern Recognition (CVPR)}, 2020, pp. 153--162.

\bibitem{goyal2017accurate}
P.~Goyal, P.~Doll{\'a}r, R.~Girshick, P.~Noordhuis, L.~Wesolowski, A.~Kyrola,
  A.~Tulloch, Y.~Jia, and K.~He, ``Accurate, large minibatch {SGD}: training
  {I}mage{N}et in 1 hour,'' \emph{arXiv preprint arXiv:1706.02677}, 2017.

\bibitem{fan2019more}
Q.~Fan, C.-F. Chen, H.~Kuehne, M.~Pistoia, and D.~Cox, ``More is less: Learning
  efficient video representations by big-little network and depthwise temporal
  aggregation,'' \emph{Advances in Neural Information Processing Systems
  (NeurIPS)}, 2019.

\bibitem{piergiovanni2019evolving}
A.~Piergiovanni, A.~Angelova, A.~Toshev, and M.~S. Ryoo, ``Evolving space-time
  neural architectures for videos,'' in \emph{International Conference on
  Computer Vision (ICCV)}, 2019, pp. 1793--1802.

\bibitem{zoph2017neural}
B.~Zoph and Q.~V. Le, ``Neural architecture search with reinforcement
  learning,'' \emph{Internation Conference on Learning Representations (ICLR)},
  2017.

\bibitem{ryoo2019assemblenet}
M.~S. Ryoo, A.~Piergiovanni, M.~Tan, and A.~Angelova, ``Assemblenet: Searching
  for multi-stream neural connectivity in video architectures,''
  \emph{Internation Conference on Learning Representations (ICLR)}, 2020.

\bibitem{zhou2018temporal}
B.~Zhou, A.~Andonian, A.~Oliva, and A.~Torralba, ``Temporal relational
  reasoning in videos,'' in \emph{European Conference on Computer Vision
  (ECCV)}, 2018, pp. 803--818.

\bibitem{li2019collaborative}
C.~Li, Q.~Zhong, D.~Xie, and S.~Pu, ``Collaborative spatiotemporal feature
  learning for video action recognition,'' in \emph{Conference on Computer
  Vision and Pattern Recognition (CVPR)}, 2019, pp. 7872--7881.

\bibitem{rumelhart1985learning}
D.~E. Rumelhart, G.~E. Hinton, and R.~J. Williams, ``Learning internal
  representations by error propagation,'' California Univ San Diego La Jolla
  Inst for Cognitive Science, Tech. Rep., 1985.

\bibitem{hochreiter1997long}
S.~Hochreiter and J.~Schmidhuber, ``Long short-term memory,'' \emph{Neural
  computation}, vol.~9, no.~8, pp. 1735--1780, 1997.

\bibitem{gers2000recurrent}
F.~A. Gers and J.~Schmidhuber, ``Recurrent nets that time and count,'' in
  \emph{International Joint Conference on Neural Networks (IJCNN)}, vol.~3,
  2000, pp. 189--194.

\bibitem{zeiler2013stochastic}
M.~D. Zeiler and R.~Fergus, ``Stochastic pooling for regularization of deep
  convolutional neural networks,'' in \emph{International Conference on
  Learning Representationsm (ICLR)}, 2013.

\bibitem{ghadiyaram2019large}
D.~Ghadiyaram, D.~Tran, and D.~Mahajan, ``Large-scale weakly-supervised
  pre-training for video action recognition,'' in \emph{Conference on Computer
  Vision and Pattern Recognition (CVPR)}, 2019.

\end{thebibliography}
}

\end{document}